\documentclass{article}

\usepackage{arxiv}

\usepackage[utf8]{inputenc} 
\usepackage[T1]{fontenc} 
\usepackage{hyperref} 
\usepackage{url} 
\usepackage{booktabs} 
\usepackage{amsfonts} 
\usepackage{nicefrac} 
\usepackage{microtype} 
\usepackage{lipsum}		
\usepackage{graphicx}
\usepackage{natbib}
\usepackage{doi}
\usepackage{tablefootnote}

\usepackage{stmaryrd} 
\usepackage{amsmath} 
\usepackage{float} 
\usepackage{arydshln} 
\usepackage{amssymb} 
\newtheorem{theorem}{Theorem}

\title{High-dimensional multi-view clustering methods}

\author{ 
	\href{https://orcid.org/0000-0000-0000-0000}{\includegraphics[scale=0.002]{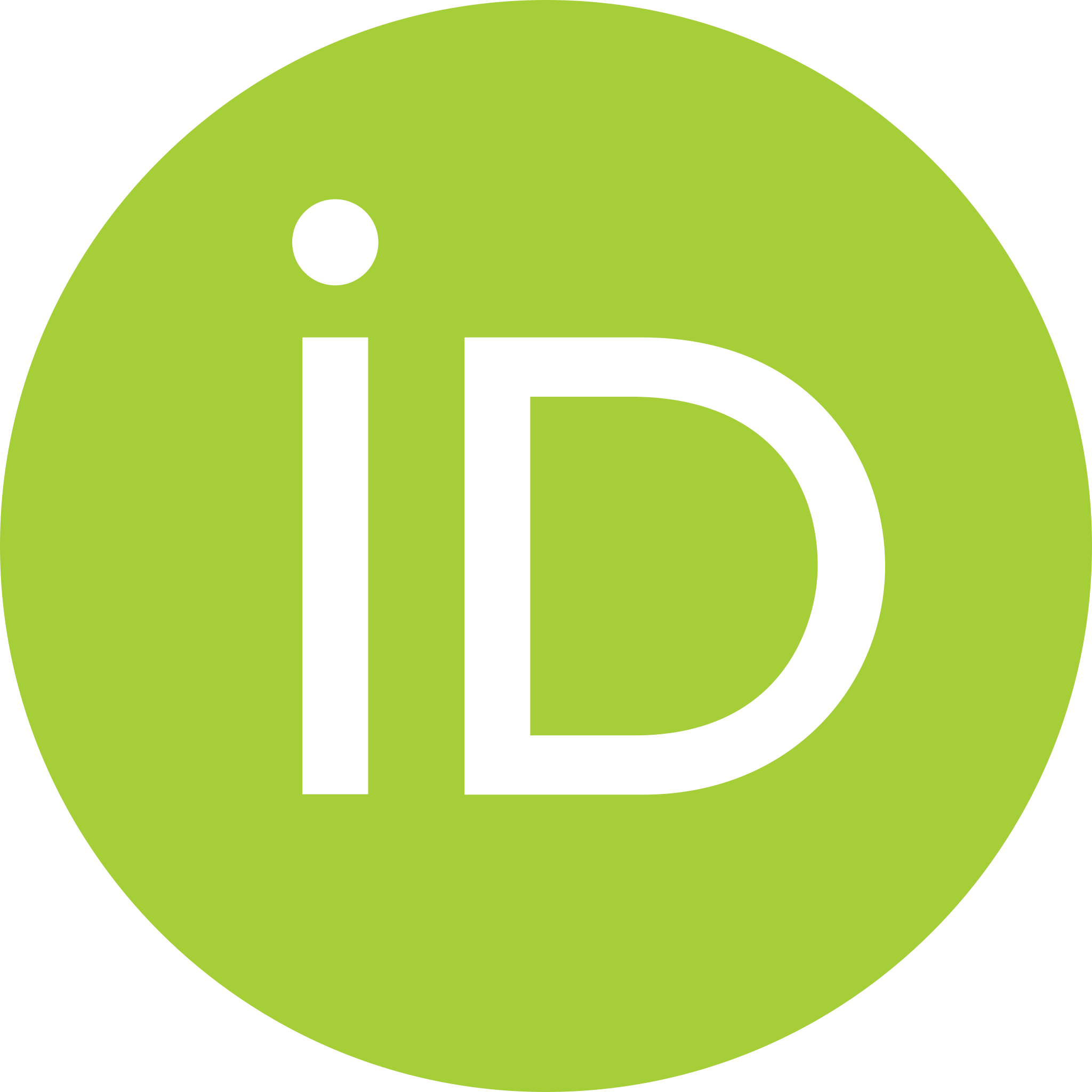}\hspace{1mm}K. Jbilou} \\
	Approximation and Numerical Analysis department\\
	ULCO University\\
	Dunkerque, France 62228 \\
	\texttt{khalide.jbilou@univ-littoral.fr} \\
 	\And
	\href{https://orcid.org/0000-0000-0000-0000}{\includegraphics[scale=0.002]{orcid.png}\hspace{1mm}A. Ratnani} \\
	Khwarizmi department\\
 Mohammed VI Polytechnic University\\
	Benguerir, Morocco 43150 \\
	\texttt{ahmed.ratnani@um6p.ma} \\
 \and
 \href{https://orcid.org/0000-0002-7560-7216}{\includegraphics[scale=0.002]{orcid.png}\hspace{1mm}A. Zahir}\\
 	Khwarizmi department\\
 Mohammed VI Polytechnic University\\
 	Benguerir, Morocco 43150 \\
 	\texttt{alaeddine.zahir@um6p.ma} \\
}

\date{\today}

\renewcommand{\shorttitle}{MVC methods}

\hypersetup{
pdftitle={A template for the arxiv style},
pdfsubject={q-bio.NC, q-bio.QM},
pdfauthor={David S.~Hippocampus, Elias D.~Striatum},
pdfkeywords={First keyword, Second keyword, More},
}

\begin{document}
\maketitle

\begin{abstract}
Multi-view clustering has been widely used in recent years in comparison to single-view clustering, for clear reasons, as it offers more insights into the data, which has brought with it some challenges, such as how to combine these views or features. Most of recent work in this field focuses mainly on tensor representation instead of treating the data as simple matrices. This permits to deal with the high-order correlation between the data which the based matrix approach struggles to capture. Accordingly, we will examine and compare these approaches, particularly in two categories, namely graph-based clustering and subspace-based clustering. We will conduct and report experiments of the main clustering methods over a benchmark datasets.
\end{abstract}


\keywords{Multi-view clustering \and Graph based Multi-view clustering \and Subspace based Multi-view clustering \and Tensor learning \and Representation learning}

\section{Introduction}
Clustering \citep{Li_2019,Yang2018,ZHAO201743} is a fundamental topic of data mining as well as machine learning, especially when there are no labels for data objects. Clustering results are often used in subsequent applications, such as community detection, recommendation, and information retrieval using features.
A feature is any individual measurable property or characteristic of an object, for example, an image is represented by different kind of features as colors, edges or texture, a document can be represented by different language. A video may be encoded in different amount of images and sound.

\noindent
The combine of these features are refereed as multi-view data, which it gave birth to a new paradigm. Although each view may have enough information by it's own, it likely provides complementary information to the other views, this leaning paradigm is humanly nature, as we also tend to learn in a similar fashion, as in many real-life problems multi-view data arise naturally. The challenge in this new paradigm, is how to combine these features for a better result than the single view.
In this paper, we focus on multi-view unsupervised learning and particularly, multi-view clustering. Note that we will focus on a more particular subject, that will be mentioned after giving the appropriate definition at the right time.

\noindent
One naive MVC technique is to use a single-view clustering algorithm on concatenated information acquired from several views. This strategy however, may fail if greater emphasis is placed on particular specific view than on others. As a result, multi view subspace clustering that learns common coefficient matrices, multi view k-means \citep{Cai_2013} and multi-kernel based MVC \citep{Guo2014} have received increased attention.

\noindent
The remainder of the paper is structured as follows: In Section \ref{sec2}, we present the notations throughout the paper which are listed as follows according to \cite{Kilmer2011}, and state some definitions that we will use throughout this review, then in Section \ref{sec3}, we will talk about the single view clustering, and in Section \ref{sec4}, we will introduce the different categories in the multi view clustering, both the matrix based as a first part, and the tensor based in the second part, experimental results are given in Section \ref{sec5}. Finally, we give concluding remarks and future possible work in the last Section \ref{sec6}.

\section{Preliminaries}\label{sec2}
This section will present some preliminaries and notations and related to the subject.
For convenience, we summarize the frequently used notations in table ~\ref{Tab:Notations}.\\
Throughout this review, we will denote the matrices are denoted by boldface capital letters (e.g $X$), an element of a matrix is denoted by $x_{i j}$. Vectors will be denoted by a boldface lowercase letters (i.e., $\mathbf{x}$), and an element of a vector is denoted by $x_{i}$.
The Frobenius norm of $X$ is denoted by $\|X\|_{F}$.
The $\ell_{2,1}$ norm of a matrix that is not necessary $X$, is defined as sum of norm-2 of its columns, which encourages the sparsity property of these columns. 
$\mathbf{I}_{n}$ denotes an identity matrix with size n.

\noindent
We denote an indirect graph with $M$ views as $G=\left(V, E_{(1)}, E_{(2)}, \cdots, E_{(V)}\right)$, where $V$ is the set of nodes and $E_{(v)} \subset V \times V$ is the set of edges from view $v$ of $G$. We denote $S_{(v)} \in \mathbb{R}^{n \times n}$ the weighted affinity matrix in the $v$-th view, and it's entries denote the pairwise affinity between nodes of $G$ in the $ v$-th view.\\
We denote by the diagonal matrix $D^{(v)}$ the degree matrix of $G$ in view $v$ where $D_{i}^{(v)}=\sum_{j=1}^{n} S_{i,j}^{(v)}$. We also denote by $L^{(v)}=D^{(v)} - A^{(v)}$ the Laplacian matrix of $G$ in the $v$-th view, if not specified otherwise.

\noindent
We will denote tensors calligraphy letters (i.e., $\mathcal{X}$). For third order tensors $\mathcal{X} \in \mathbb{R}^{N_{1} \times N_{2} \times N_{3}}$, the i-th horizontal slice, j-th lateral slice and k-th frontal slice are defined by $\mathcal{X}(i,:,:,),\mathcal{X}(:,j,:)$ and $\mathcal{X}(:,:,k)$ respectively.\\
We can transform a tensor to a matrix in mode k, named the mode-k Matricization of a tensor $\mathcal{X}$ denoted, $\mathcal{X}_{(k)} \in \mathbb{R}^{N_{k} \times J}$
where $J=\Pi_{q=1, q \neq k}^{3} N_{q}$\\
We can also define the product of a tensor by a matrix $U$ in mode k, named the k-mode (matrix) of a tensor denoted $\times{ }_{k}$: 
$$\mathcal{Y}=\mathcal{X} \times{ }_{k} U \Leftrightarrow Y_{(k)}=U X_{(k)}.$$
We define some tensor decomposition, namely the CP, tucker and T-svd decomposition for third order tensors:
a) The Tucker decomposition is a form of higher-order PCA. It decomposes a tensor
into a core tensor multiplied (or transformed) by a matrix along each mode:
\begin{equation}
\mathcal{Z} \approx \mathcal{G} \times_{1} U_1 \times_{2} U_2 \times_{3} U_3,
\end{equation}
where the tensor $\mathcal{G}$ is called, the core tensor. The matrices $U_k$, $k=1,\ldots,3$ are the factor matrices (which are usually orthogonal) and can be thought of as the principal components in each mode.\\
b) The CANDECOMP / PARAFAC (CP) factorization by:
\begin{equation}
\mathcal{Z} \approx \llbracket Z^{(1)},Z^{(2)}, Z^{(3)} \rrbracket = \sum_{r=1}^{R} \mathbf{z}_{r}^{(1)} \circ \mathbf{z}_{r}^{(2)} \circ \mathbf{z}_{r}^{(3)},
\end{equation}
with $Z^{(k)}=\left[\mathbf{z}_{1}^{(k)},\mathbf{z}_{2}^{(k)} \cdots, \mathbf{z}_{R}^{(k)}\right]$ are factor matrices, $R$ is the number of factors, and $\llbracket \cdot \rrbracket$ is used for shorthand, and $\circ$ is the outer product.\\
These two decomposition can be defined for higher order tensors, the next decomposition (T-SVD) is reserved for the 3 order tensors.\\
c) The T-SVD decomposition: Before defining it, or even the t-product. We start by defining the Discrete Fourier Transformation for vectors before showing The DFT on tensors.

The Discrete Fourier Transformation (DFT) is essential in the tensor-tensor product. We provide some background information and notations here. The DFT matrix $F_n$ has the following structure:
$$F_n=[w^{(i-1)(j-1)}],\; i,j=1,\ldots,n,$$
where $w=e^{\frac{2\pi i}{n}}$ is the primitive n-root of 1.\\
The matrix $F_n$ verifies $F_n F_n^{*}=n I_n$, which means that $F_n^{-1}=\frac{1}{n} F_n^{*} $. The symbol $^{*}$ represents the conjugate transpose. 
The DFT of a vector $\mathbf{v}$ denoted by $\widetilde{\mathbf{v}}=fft(\mathbf{v})$ is given by 
\begin{equation}
\widetilde{\mathbf{v}}=F_n \mathbf{v} \in \mathbb{C}^n.
\label{eq:Dft_vector}
\end{equation}
Denote the circulant matrix of $\mathbf{v}$ as
\begin{equation*}
\operatorname{circ}(\mathbf{v})=\left[\begin{array}{cccc}
v_1 & v_n & \cdots & v_2 \\
v_2 & v_1 & \cdots & v_3 \\
\vdots & \vdots & \ddots & \vdots \\
v_n & v_{n-1} & \cdots & v_1
\end{array}\right] \in \mathbb{R}^{n \times n}.
\end{equation*}
It can be diagonalized by the DFT matrix, i.e.,
\begin{equation}
F_n \operatorname{circ}(\mathbf{v}) F_n^{-1} =diag(\widetilde{\mathbf{v}}). 
\label{eq:DFT_diag_vect}
\end{equation}
For any vector $\mathbf{w} \in \mathbb{R}^n$, the associated $\widetilde{\mathbf{w}}=F_n \mathbf{w}$ satisfies :
\begin{equation}
 \widetilde{w}_1 \in \mathbb{R}, \quad conj(\widetilde{w_i})=\widetilde{w}_{n-i+2}, \text{for} \; i=2,3 \dots \lfloor\frac{n+1}{2}\rfloor 
 \label{eq:DFT_ifft_vect}
\end{equation}
Now we consider the DFT for tensors,
We denote $\widetilde{\mathcal{X}}$ the discrete Fast Fourier transform (FFT) of $\mathcal{X}$ along the third dimension i.e., $\widetilde{\mathcal{X}}=f f t(\mathcal{X},[\,], 3)$. Similarly, the tensor can be obtained back by the inverse FFT (IFFT) along the third dimension $\mathcal{X}=i f f t(\widetilde{\mathcal{X}},[\,], 3)$. For a tensor $\mathcal{X}$ whose frontal slices are $X^{(1)},\ldots, X^{\left(N_{3}\right)}$, 
we define the block diagonal matrix, the block curculant matrix, an unfolding operator and it's inverse, unfold as:
$$
\operatorname{bcirc}(\mathcal{X})=\left[\begin{array}{cccc}
X^{(1)} & X^{\left(N_{3}\right)} & \cdots & X^{(2)} \\
X^{(2)} & X^{(1)} & \cdots & X^{(3)} \\
\vdots & \vdots & \ddots & \vdots \\
X^{\left(N_{3}\right)} & X^{\left(N_{3}-1\right)} & \cdots & X^{(1)} 
\end{array}\right],
$$
$$\operatorname{bdiag}
( \mathcal { X })
=\left[\begin{array}{llll}
X^{(1)} & & & \\
& X^{(2)} & & \\
& & \ddots & \\
& & & X^{\left(N_{3}\right)}
\end{array}\right],$$
$$
\operatorname{unfold}(\mathcal{X})=\left[\begin{array}{c}
X^{(1)} \\
X^{(2)} \\
\vdots \\
X^{\left(N_{3}\right)}
\end{array}\right],
\text { fold }(\operatorname{unfold}(\mathcal{X}))=\mathcal{X}.
$$
The block circulant matrix can be block-diagonalized as follows:
\begin{equation}
\left(F_{N_3} \otimes I_{N_1}\right) \cdot \operatorname{bcirc}(\mathcal{X}) \cdot \left(F_{n_3}^{-1} \otimes I_{N_2}\right)=\operatorname{bdiag}
( \mathcal { \widetilde{X})},
\label{eq:DFT_diag}
\end{equation}
where $\otimes$ denotes the Kronecker product. By using the same lemma as before we have
\begin{equation}
\left\{\begin{array}{l}
\widetilde{X}^{(1)} \in \mathbb{R}^{N_1 \times N_2} \\
\operatorname{conj}\left(\widetilde{X}^{(i)}\right)=\widetilde{X}^{\left(N_3-i+2\right)}, i=2, \cdots,\left\lfloor\frac{N_3+1}{2}\right\rfloor.
\end{array}\right.
\label{eq:DFT_ifft}
\end{equation}
In contrast, for any given $\widetilde{\mathcal{X}} \in \mathbb{C}^{N_1 \times N_2 \times N_3}$ satisfying \eqref{eq:DFT_ifft}, there exists a real tensor $\mathcal{X} \in \mathbb{R}^{N_1 \times N_2 \times N_3}$ such that \eqref{eq:DFT_diag} is satisfied.\\
The t-product $\mathcal{X} * \mathcal{B}$, with $\mathcal{B} \in \mathbb{R}^{N_{2} \times L \times N_{3}}$ is a tensor of size $N_{1} \times L \times N_{3}$ defined by:
\begin{equation}
\mathcal{X} * \mathcal{B}=\text { fold }(\operatorname{bcirc}(\mathcal{X}) \cdot \text { unfold }(\mathcal { B })).
\end{equation}
The t-product is also called the circular convolution operation. It is shown that
$$\mathcal{C}=\mathcal{X} * \mathcal{B} \iff \widetilde{C}=\widetilde{X} \widetilde{B},$$
which suggests an efficient way to calculate the t-product based on the FFT.\\
The transpose of $\mathcal{X}$ denoted by $\mathcal{X}^{T} \in \mathbb{R}^{N_{2} \times N_{1} \times N_{3}}$ is obtained by transposing each of the frontal slices and then reversing the order of transposed frontal slices 2 through $N_{3}$.\\
The identity tensor $\mathcal{I}$ is a tensor whose first frontal slice is the identity matrix and the remaining frontal slices has zero entries.\\
A tensor is orthogonal if it satisfies the relationship:
$$\mathcal{X}^{T} * \mathcal{X}=\mathcal{X} * \mathcal{X}^{T}=\mathcal{I}.$$
A tensor is called, f-diagonal if each of its frontal slices in the fourier domain is a diagonal matrix.\\
The T-SVD decomposition theorem \citep{Kilmer2011,Kilmer2013,Hao2013} states that $\mathcal{X}$ can be factorized as: 
\begin{equation}
\mathcal{X}=\mathcal{U} * \mathcal{S} * \mathcal{V}^{*}, 
\end{equation}
where $\mathcal{U} \in \mathbb{R}^{N_{1} \times N_{1} \times N_{3}}, \mathcal{V} \in \mathbb{R}^{N_{2} \times N_{2} \times N_{3}}$ are orthogonal, and $\mathcal{S} \in \mathbb{R}^{N_{1} \times N_{2} \times N_{3}}$ is an f-diagonal tensor. 
It can also be written as 
$$\mathcal{X}=\sum_{i=1}^{\min \left(N_{1}, N_{2}\right)} \mathcal{U}(:, i,:) * \mathcal{S}(i, i,:) * \mathcal{V}(:, i,:)^{\mathrm{T}}.$$
We also have a truncated T-SVD representation that provide an optimal approximation in the same way as the truncated matrix SVD: for $k<\min \left(N_{1}, N_{2}\right)$, we define $\mathcal{X}_{k}= \sum_{i=1}^{k} \mathcal{U}(:, i,:) * \mathcal{S}(i, i,:) * \mathcal{V}(:, i,:)^{\mathrm{T}}$, then we have the following result:
$$\mathcal{X}_{k}=\underset{\widetilde{\mathcal{X}} \in \mathbb{M}}{\operatorname{argmin}}\|\mathcal{X}-\widetilde{\mathcal{X}}\|_{F},$$
where $\mathbb{M}=\left\{\mathcal{C}=\mathcal{X} * \mathcal{Y} \mid \mathcal{X} \in \mathbb{R}^{N_{1} \times k \times N_{3}}, \mathcal{Y} \in \mathbb{R}^{k \times N_{2} \times N_{3}}\right\}$
which gives a best low rank approximation to the matrix in terms of the Frobenius norm under rank k constraint.\\
We define the T-SVD based tensor nuclear norm (T-TNN) \citep{Zhang2014}:
$$
\|\mathcal{X}\|_{\circledast}=\sum_{i=1}^{N_{3}}\left\|\widetilde{X}^{(i)}\right\|_{*},
$$
where $\left\|.\right\|_{*}$ is the nuclear norm, which is exactly the sum of all its singular values. It is proven to be a valid norm and the tightest convex relaxation to $l_1$ norm of the tensor multi-rank.\\ 
We can further define a more generalization of this norm, called the weighted tensor nuclear norm:
$$
\|\mathcal{X}\|_{\omega, \circledast}=\sum_{i=1}^{N_3}\left\|\widetilde{X}^{(i)}\right\|_{\omega, *}=\sum_{i=1}^{N_3} \sum_{j=1}^{\min \left(N_{1}, N_{2}\right)} \omega_{j} * \sigma_{j}\left(\widetilde{X}^{(i)}\right) \text {, }
$$
where, $\sigma_{j}\left(\widetilde{X}^{(i)}\right)$ denotes the $j$ largest singular value of $\widetilde{X}^{(i)}$.\\
We also define the Shatter p-norm, for $0<p \leqslant 1$:
$$
\|\mathcal{X}\|_{S p}^{p} = \sum_{i=1}^{N_3}\left\|\widetilde{\mathcal{X}}^{(i)}\right\|_{Sp}^{p}= \sum_{i=1}^{N_3} \sum_{j=1}^{\min \left(N_{1}, N_{2}\right)} \sigma_{j}\left(\widetilde{\mathcal{X}}^{(i)}\right)^{p}.
$$
A norm that will be presented is the t-Gamma tensor quasi-norms presented by \citep{Wu2020} defined as:
$$
\|\mathcal{X}\|_{t-\gamma}=\frac{1}{N_3} \sum_{n_3=1}^{N_3}\left\|\widetilde{X}^{(n_3)}\right\|_\gamma,
$$
where $\|.\|_\gamma$ represents the Gamma quasi-norm for matrices \citep{Wu2020} , for $Z \in \mathbb{R}^{N_1 \times N_2}$
$$\|Z\|_\gamma=\sum_{i=1}^{\min(N_1, N_2)} \frac{(1+\gamma) \sigma_i(Z)}{\gamma+\sigma_i(Z)}, \gamma>0.$$
The table \ref{Tab:Notations} summarizes the different important notations that will be used in this work, for convenience, we add a dashline that separates the notations related to tensors.
\begin{table}[ht]
\begin{center}
\label{Tab:Notations}
\begin{tabular}{l|l|l|l}
\hline$x$,$\mathbf{x}$ & Scalar, vector. & $X$, $\mathcal{X}$ & Matrix, tensor \\
$\|X\|_{F}$ & $\|X\|_{F}=\sqrt{\sum_{i j} X_{i j}^{2}}. $ & $\|X\|_{*}$ & Nuclear norm. \\
$\operatorname{tr}(X)$ & The trace function. & $\|X\|_{2,1}$ & $\|X\|_{2,1}=\sum_{j}\|X(:, j)\|_{2}. $ \\
$\operatorname{Diag}(X)$ & The diagonal entries. & $\|X\|_\gamma$ & Gamma quasi-norm.\\
$X^{*}$ & The conjugate transpose. & & \\ 
\hdashline
$\mathcal{X}_{i j k}$ & The $(i, j, k)$-th entry of $\mathcal{X}. $ & $\|\mathcal{X}\|_{F}$ & $\|\mathcal{X}\|_{F}=\sqrt{\sum_{i j k}\mathcal{X}_{i j k}^{2}}$ \\
$\mathcal{X}(i,:,:)$ & The $i$-th horizontal slice of $\mathcal{X}. $ & $\|\mathcal{X}\|_{1}$ & $\|\mathcal{X}\|_{1}=\sum_{i j k} \mid \mathcal{X}_{i j k} \mid . $ \\
$\mathcal{X}(:,i,:)$ & The $i$-th lateral slice of $\mathcal{X}. $ & $\|\mathcal{X}\|_{2,1}$ & $\|\mathcal{X}\|_{2,1}=\sum_{i, j}\|\mathcal{X}(i, j,:)\|_{2}. $ \\
$X^{(i)}$ & The $i$-th frontal slice of $\mathcal{X}. $ & $\|\mathcal{X}\|_{\omega, \circledast}$ & \footnotesize{Weighted based tensor nuclear norm.} \\
$\widetilde{\mathcal{X}}$ & $\widetilde{\mathcal{X}}=\operatorname{fft}(\mathcal{X},[], 3). $ & $\|\mathcal{X}\|_{\circledast}$ & T-SVD based tensor nuclear norm. \\
$\mathcal{X}_{(i)}$ & Mode-i matricization of $\mathcal{X}. $ & $\|\mathcal{X}\|_{S p}^p$ & Shatten p-norm. \\
$\mathcal{X}^{T}$ & The transpose of $\mathcal{X}. $ & $\|\mathcal{X}\|_{FF1}$ & $\|\mathcal{X}\|_{FF1}=\sum_{i}\|\mathcal{X}(i,:,:)\|_{2} $. \\
$\widetilde{\mathcal{X}}$ & The Fast Fourier Transform of $\mathcal{X}$. & $\|\mathcal{X}\|_{t-\gamma}$ & T-Gamma tensor quasi-norms. \\
\hline
\end{tabular}
\end{center}
\end{table}

Next, we will show some clustering methods. There are roughly two categories of clustering that depend on the number of views: Single view clustering, and the more general framework, the multi-view clustering.

\section{Single view clustering}\label{sec3}
Single view clustering is performing the clustering task using only single view feature. It fails to explore the correlations among the other features from different sources.\\
Given a feature matrix $X \in \mathbb{R}^{d \times n}$, where $n$ represents the number of samples, and $d$ is the dimension feature. 
Two categories that are going to be presented: Graph based and subspace.

\subsection{Single view graph based clustering:}
In single view graph based clustering, each node or a vertex of a graph corresponds to an object and each edge represents a relationship between two objects. The general methods that are in this category, follow the following steps to partitions the $n$ data points into $c$ clusters:

\begin{enumerate}
\item Construct a data graph matrix $S \in \mathbb{R}^{n \times n}$, where each entry in $S$ represents the similarity between the points.
\item Compute the graph Laplacian matrix $L_{S}=D_{S}-\frac{S^{T}+.S}{2}$, where $D_{\mathrm{S}}$ is a diagonal matrix: $d_{i,i}=\sum_{j} \dfrac{s_{i j}+s_{j i}}{2}$.
\item Compute the embedding matrix $F \in \mathbb{R}^{n \times c}$ by solving the following problem \\$\min_{F \in \mathbb{R}^{n \times c}} \operatorname{Tr}\left(F^{T} L_{S} F\right)=\min_{F \in \mathbb{R}^{n \times c}} \frac{1}{2} \sum_{i,j} s_{i j}||\mathbf{f}_i-\mathbf{f}_j||_2$.
\item Clustering F into $c$ groups with an additional clustering algorithm (e.g., K-means).
\end{enumerate}

\subsection{Single view subspace clustering} 
The subspace learning approach is built on the assumption that all the views are generated from a latent subspace. Without loss of generality, we can write the data matrix $X$ in the matrix form $X= X Z+E$, where $Z \in \mathbb{R}^{n \times n}$ is the representation matrix, $E \in \mathbb{R}^{d \times n}$ is the error matrix. 
The general method that falls in this category aims to solve the following optimization problem
\begin{equation}
\min _{Z, E} \Psi(Z)+\lambda \varphi(E) \; \text{ s.t } X=X Z+E, 
\end{equation}
where $\Psi(Z)$ is the regularization which imposes desired property on $Z \in \mathbb{R}^{n \times n}, 
\varphi(E)$ is the loss function to remove noise and $\lambda>0$ is a balanced parameter that is controlled by the user. 
It can also be added other constraints, for example $Diag(Z)=0$, which enforces that each data point can only be represented as a combination of the remaining points except itself. 
After getting the representation matrix $Z$, we compute the affinity matrix and further employ it as the input of the spectral clustering algorithm for achieving the final clustering result.
\\
As such, Low-Rank Representation method (LRR) \citep{Liu_2010} introduces the low rank regularization to subspace clustering by putting the regularization as the nuclear norm, and the loss function as the $l_{2-1}$ norm. On the other hand, Sparse Subspace Clustering (SSC) \citep{Ehsan}, represents each data point as a sparse linear combination of other points and captures the data local structure.
SSC is based on self representation property i.e., data matrix stands for a dictionary. 
According to the experimental results in \citep{Wang2013}, LRR misclassifies different data points than SSC. As a result, in order to capture both the global and local structure of the data, low-rank and sparsity requirements must be combined \citep{Wang2013}.
\section{Multi-view clustering}\label{sec4}
Multi-view clustering tries to benefit from all these views, to perform the clustering task. The success of the multi-view leanings is based on two principles: the complementary principal and the consensus principal. Specifically, the complementary principle, which means that a view can contain information that the others do not have. The consensus principle aims to maximize the agreement on the different views, which makes the MVC has better robustness partition and accuracy than the single view clustering. The core challenge of MVC is how to integrate all the different features to obtain a reliable affinity matrix.

\begin{figure}[ht]
\centering
\includegraphics[scale= 0.3]{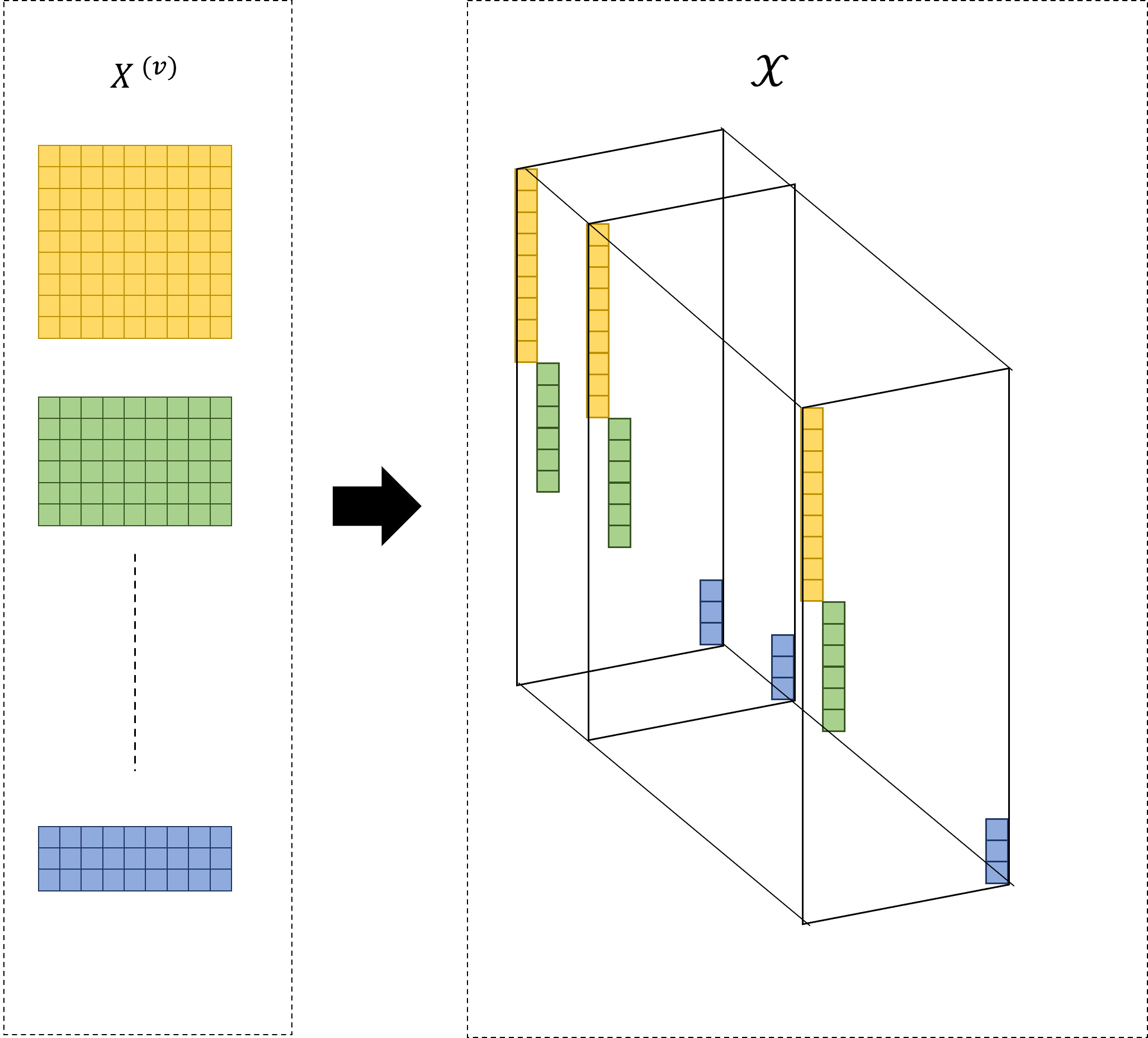}
\caption{Constructing a third order tensor from the matrices in $ \mathbb{R}^{d_{v} \times n}$}
\label{fig:Tensor_construction}
\end{figure}

\noindent
MVC can also be categorized into roughly 5 categories, co-training style algorithms, multi-kernel learning, multi-view graph clustering, multi-view subspace clustering and multi-task multi-view clustering. 
Co-training based methods search for the results that agree across different views using a co-training strategy, by bootstraps the partitions of different views by using the prior or learned knowledge from one another. The method relies on conditional independence.
Co-regularization based methods jointly regularize the hypotheses to explore the complementary information.
Multiple Kernel Learning (MKL) methods usually combine different kernels by adding them equally or learning the combination weights either linearly or non-linearly. The method have high computational complexity.\\
We will first discuss the data construction:\\
Given a multi-view data $\{ X^{(v)} \}_{v=1,\dots,V}$, where $X^{(v)}=[\mathbf{x}_{1}^{(v)},\mathbf{x}_{2}^{(v)} \ldots, \mathbf{x}_{n}^{(v)}] \in \mathbb{R}^{d_{v} \times n}$ is the data matrix of the $v$-th view, $d_{v}$ is the dimension of the $v$-th view, $n$ is the number of data points and $V$ is number of views. Although the feature dimension is variable, we can always construct a third order tensor $\mathcal{X}$, first we create a block diagonal matrix $B_i \in \mathbb{R}^{D \times V}$ with $D=\sum_v d_{v}$ as such:
$$B_i=\left[\begin{array}{llll}
X^{(1)}(:,i) & & & \\
& X^{(2)}(:,i) & & \\
& & \ddots & \\
& & & X^{(V)}(:,i)
\end{array}\right]$$ then twisting it to obtain $\mathcal{B}_i \in \mathbb{R}^{D \times 1 \times V}$, and finally, construct the tensor $\mathcal{X} \in \mathbb{R}^{D \times n \times V}$ by concatenating the n tensors $\mathcal{B}_i$ along the second mode as in figure \ref{fig:Tensor_construction}.

\noindent
By following these steps, we can rid of the variable dimension, while preserving the structure of the data.
\subsection{Multi-view graph clustering}
Multi-view graph clustering aims to find a fusion graph of all views, then apply the classical clustering methods on the fusion graph, as spectral clustering or graph-cut algorithms to produces the final clustering. 
\begin{figure}
\centering
\includegraphics[scale= 0.3]{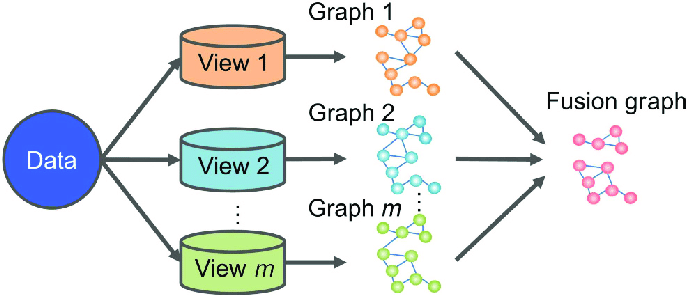}
\caption{The general method of multi-view graph-based clustering \cite{Yang2018}}
\end{figure}
Spectral clustering relies on computing the eigenvectors of the graph Laplacian matrix.\\
One of the most representative spectral clustering methods is the co-regularized spectral clustering. It performs standard spectral clustering on each view to get the corresponding indicator matrix, and then gets the common indicator matrix by minimizing disagreement between indicator matrices. However, it treats each indicator matrix equally, which is not ideal in real applications.

\noindent
The challenge is how to assign an appropriate weight to each view, even though some method \cite{Chaudhuri2009} don't consider specific these different views. This strategy neglects the importance of different graphs and may suffer when an unreliable graph is added to, other methods (e.g., \cite{Xia2010}) add new hyper parameters, but it's effect on the clustering is cannot be underestimated. Other methods learn the weight automatically, which eliminates the need of any additional hyper parameter, as we will present some in the following.
\noindent
The method used in Auto-weighted Multiple Graph Learning (AMGL) \cite{Nie_2016}, supposes that the normalized Laplacian matrices $L_S^{(v)}$ are constructed, and then solve the following problem:
\begin{equation}
\min _{F \in \mathbb{R}^{n \times c} F^{T} F=\mathbf{I}}\sum_{v=1}^{V} \sqrt{\operatorname{Tr}\left(F^{T} L_{S}^{(v)} F\right)}.
\end{equation}
which is solved by fixing the parameters $\alpha$ then solves: 
\begin{equation*}
\min _{F \in \mathbb{R}^{n \times c} F^{T} F=\mathbf{I}}\sum_{v=1}^{V} \alpha^{(v)}\operatorname{Tr}\left(F^{T} L_{S}^{(v)} F\right).
\end{equation*}
Then, the parameters is updated using the formula:
$$\alpha^{(v)} = \dfrac{1}{\operatorname{Tr}\left(F^{T} L_{S}^{(v)} F\right)}.$$
The method can also be used for semi supervised clustering, but it is not in the scope of this review. In AMGL method, there are no hyper parameters, it can learn both the weight of each view and also the embedding matrix $F$, and then we treat each row of $F$ as a new representation of each data point and compute the clustering labels by using k-means algorithm.

After merging the view, some methods modify the traditional spectral clustering framework, they don't need to search the uniform similarity matrix, but instead, try to learn an indicator matrix, that is used directly for clustering. 
One can also impose a constraint Laplacian rank, which guarantees that we have c clusters (components in the language of graphs), as stated by the following theorem.
\begin{theorem}\citep{Marsden2013}
The multiplicity c of the eigenvalue 0 of the Laplacian matrix $L_S$ is equal to the number of connected components in the graph with the similarity matrix $S$.
\end{theorem}
\noindent
In \citep{Nie_2017}, a Parameter-weighted Multi view Clustering (PwMC) method has been proposed. It is based on the generalization of the CLR method: It learns a new similarity matrix from the old similarities of the v-th view $S^{(v)}$, computed from the original data:
\begin{equation}
\begin{aligned}
\min _{w_{(v)}, U} & \sum_{v=1}^{V} w^{(v)}\left\|U-S^{(v)}\right\|_{F}^{2}+\gamma\|\mathbf{w}\|_{2}^{2} \\
\text { s.t. } & w_{(v)} \geq 0, \mathbf{1}_{v}^{T} \mathbf{w} =1, u_{i j} \geq 0, \mathbf{1}^{T} \mathbf{u}=1 \\ & \operatorname{rank}\left(L_{U}\right)=n-c.
\end{aligned}
\label{eq:PwMC}
\end{equation}
The rank constraint is hard to tackle, since the diagonal matrix $D_S$ depends also on $S$. As the Laplacian is positive semi-definite, it's eigenvalues are positive, which makes the constraint rank:
$\operatorname{rank}\left(L_{U}\right)=n-c$ can be ensured by $\sum_{i=1}^{c} \sigma_{i}\left(L_{U}\right)=0$, where $\sigma_{i}(V)$ represents the i-th smallest eigenvalue of a matrix $M$. Thus we can replace the constraint, by adding $\lambda \sum_{i=1}^{c} \sigma_{i}\left(L_{U}\right)$ to the objective function, with $\lambda$ large enough. In addition, we can further make the objective function easier to solve, since we have the following theorem:
\begin{theorem}\citep{Fan1949}
	We have the following result 
$$\sum_{i=1}^{c} \sigma_{i}\left(L_{S}\right)=\min _{F \in \mathbb{R}^{n \times c}, F^{T} F=\mathbf{I}} \operatorname{Tr}\left(F^{T} L_{S} F\right).$$
\end{theorem}
\noindent
The method depends on the hyper parameter $\gamma$, which affects the clustering accuracy since it is confirmed to be dataset-related. A new method proposed named Self-weighted Multi-view Clustering (SwMC) \citep{Jing2017}, deals with this problem by removing completely $\gamma\|w\|_{2}^{2}$, from the equation \eqref{eq:PwMC}. 

\noindent
As SwMc, supposes that the similarity matrix of each view is given, or calculated separately by k-means for example, Some approaches bypass this stage since the performance is significantly dependent on the view graphs that are generated. Due to the complex distribution of data in real-world applications, it is challenging to manually create a suitable similarity graph. When the graph quality is inadequate, the performance of these approaches drops dramatically. This decreases the algorithm flexibility. One solution is to calculate the similarity matrix directly from the data points, which is found more reliable and noise reducing, in addition, it calculates the weights automatically and eliminates the post processing step as well, the method named Multi-View Clustering and Semi-Supervised Classification with Adaptive Neighbours (MLAN) \citep{Nie_2017} solves the following problem:
\begin{equation}
\label{eq: MLAN}
\begin{aligned}
&\min _{S} \sum_{v=1}^{V} \sqrt{\operatorname{tr}\left(X^{(v)} L_{S} X^{(v)}\right)}+\alpha\|S\|_{F}^{2} \\
&\text { s.t. } \mathbf{s}_{i}^{T} \mathbf{1}=1,0 \leq s_{i j} \leq 1, \quad \operatorname{rank}\left(L_{S}\right)=n-c, 
\end{aligned}
\end{equation}
Where the following relationship is used instead in \citep{Nie_2017} :
\begin{equation}
\label{eq: Trace_Lap}
\sum_{v=1}^{V} \operatorname{tr}\left(X^{(v)} L_{S} X^{(v)}\right)=\sum_{v=1}^{V} \sum_{i, j}\left\|\mathbf{x}_{i}^{(v)}-\mathbf{x}_{j}^{(v)}\right\|_{2}^{2} s_{i j}.
\end{equation}
This part \eqref{eq: Trace_Lap} is usually added to give an affinity matrix the following property: High intra-cluster similarity and low inter-cluster similarity: For each data point $x_i$, it belongs to one of the c classes, and can be connected by any other data point $x_j$ with a probability $s_{ij}$, this probability can be seen as the similarity between the points, thus, a large probability should be given to the closer sampler, in other words, a small distance $\left\|\mathbf{x}_{i}^{(v)}- \mathbf{x}_{j}^{(v)}\right\|_{2}^{2}$, which justifies the first part of the equation. In other words, the multi-view graph embedding tries to preserve the local structure of the graphs, and graphs with similar local structure tend to be close to each other in the original multi-view feature space.\\
MLAN only learns a global graph for all views (without building a graph for each view), thus a method named Graph-based Multi-view Clustering (GMC) \cite{Wang2020} tries to find both while guarding the same objective:
\begin{equation}
\begin{aligned}
\min _{S^{(v)}, U} &\sum_{v=1}^{V} \left( \operatorname{tr}\left( X^{(v)} L_{S} X^{(v)} \right)+\beta \left\|S^{(v)}\right\|_{2,1}^{2} \right) \\
&+\sum_{v=1}^{V} w_{v}\left\|U-S^{(v)}\right\|_{F}^{2}+2 \lambda \operatorname{Tr}\left(F^{T} L_{U} F\right) \\
&\text { s.t. } s_{i i}^{(v)}=0, s_{i j}^{(v)} \geq 0, \mathbf{1}^{T} \mathbf{s}_{i}^{(v)}=1, \\
&u_{i j} \geq 0, \mathbf{1}^{T} \mathbf{u}_{i}=1, F^{T} F=\mathbf{I}. 
\end{aligned}
\end{equation}
As most methods used only one similarity graph. A recent work, \citep{Rong_2021} created a multi-metric similarity graph refinement and fusion strategy for MVC in which several similarity graphs were built using different metrics and an informative unified graph was obtained by adaptively fusing these similarity graphs.

\subsection{Multi-view subspace clustering}
Multi-view subspace clustering attempts to discover a latent space from all the different views in order to deal with high-dimensional data, and then applies a traditional clustering approach to this subspace.
\begin{figure}
\centering
\includegraphics[scale= 0.72]{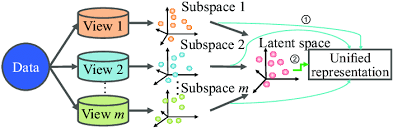}
\caption{The general method of multi-view subspace clustering \cite{Yang2018}}
\end{figure}
It normally consists of the following steps:
\begin{enumerate}
\item Learn the latent matrices or tensors using various learning methods.
\item Compute the affinity (similarity) matrix by averaging latent matrices or tensors.
\item Apply spectral clustering on the similarity matrix.
\end{enumerate}
Existing multi-view clustering algorithms may be divided into two types: Matrix-based and tensor-based representation learning.\\
Matrix theory, which organizes data in matrix form, is central to matrix-oriented approaches. It is discovered that this stream has a common flaw. It solves the problem by simply merging the views without taking into consideration the high-order correlations between these views at the same time, which is why the second stream (tensor-based representation), which has already received a lot of attention, will be the subject of this discussion.
We can generalize certain single-view methods to multi-view ways by addressing the following problem:
\begin{equation}
\begin{aligned}
\min_{Z^{(v)}, E^{(v)}} & \sum_{v=1}^{V} \left( \Psi(Z^{(v)})+\lambda \varphi(E^{(v)}) \right) \\
\text{ s.t } & X^{(v)}=X^{(v)} Z^{(v)}+E^{(v)}.
\end{aligned}
\end{equation}
After obtaining these multiple learned subspace representation $\left\{Z^{(v)}\right\}_{v=1}^{V}$, the final affinity matrix is calculated by combining all subspace representation of each view: 
\begin{equation}
S=\dfrac{1}{2V} \sum_{v=1}^{V} \left(\mid {Z}^{(v)} \mid + \mid {Z}^{(v)^{\mathrm{T}}}\mid\right).
\end{equation}
Finally, using the combined affinity matrix $S$, a spectral clustering approach or k-means is performed; however, this formulation ignores the link between subspaces and treats each subspace representation individually.

\noindent
We get multiple subspace clustering methods by using alternative norms for the loss function and regularization term, to mention a few.

\noindent
In the method named consistency-specificity multi-view subspace clustering (CSMSC) by \cite{Luo2018}, they divided the learned subspace representation $Z^{(v)}$ into a learned consistent $C$ that is shared across the views, and specific self-representation matrices $D^{(v)}$ with data under different views, as for the regularization function, the nuclear norm is chosen for $C$, to guarantee the low rank property in order to excavate more shared information among different views, and the $l_2$ norm for $D^{(v)}$ for the connectedness property, and the $l_{21}$ norm for the loss function:
\begin{equation}
\begin{aligned}
\min_{C, D^{(v)}, E^{(v)}} &\lambda_1\|C\|_{*}+\lambda_2 \sum_{v=1}^{V} \left( \left\|D^{(v)}\right\|_{2}^{2}+\left\|E^{(v)}\right\|_{2,1} \right)\\
\text { s.t. } & X^{(v)}=X^{(v)}\left(C+D^{(v)}\right)+E^{(v)}.
\end{aligned}
\end{equation}
In this case, the learned affinity matrix is calculated differently:
\begin{equation}
S=\frac{\mid C \mid +\mid C^{T}}{2} \mid +\frac{1}{2V} \sum_{v=1}^{V} \mid D^{(v)}\mid+\mid D^{(v)}\mid^{T}.
\end{equation}
Instead of the classical independence tests such as Spearmans rho and Kendalls tau that can only discover linear relationships. In the paper diversity-induced Multi-view Subspace Clustering (DiMSC), \cite{Cao_2015} used The Hilbert Schmidt Independence Criterion (HSIC) (the empirical version) as a diversity term in to recover the links between distinct subspace representations, and to enhance the complementary between the multi-view representations.

\subsection{Tensorized MVC}

\subsubsection{Tensorized subspace based MVC}
\begin{figure}[ht]
\centering
\includegraphics[scale= 0.72]{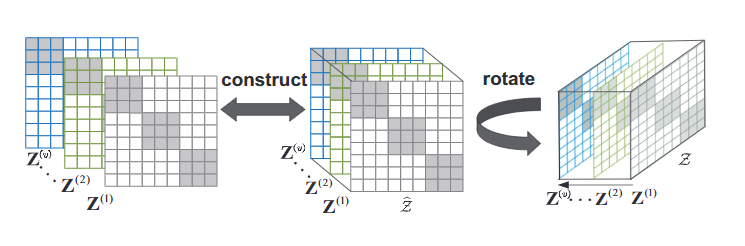}
\caption{Tensor construction and rotation it}
\label{fig:Constrcut_rotate_tensor}
\end{figure}

As previously stated, treating data as matrices rather than as a full tensor implies that the relationship between subspace is neglected, and the tensor-based problem often solves the following problem:
\begin{equation}
\label{eq:Subspace tensor MVC}
\begin{aligned}
\min_{\mathcal{Z}, \mathcal{E}} & \psi(\mathcal{Z}) + \lambda \varphi(\mathcal{E}) \\
\text{S.t. } & X^{(v)}=X^{(v)} Z^{(v)}+E^{(v)}\\
&\mathcal{Z}=\Phi_1\left(Z^{(1)}, Z^{(2)}, \ldots, Z^{(V)}\right) \\
&\mathcal{E}=\Phi_2\left(E^{(1)}, E^{(2)}, \ldots, E^{()}\right),
\end{aligned}
\end{equation}
where $\Phi$ is a suitable regularized tensor norm, $\varphi$ is a valid tensor norm for noise cancellation or sparsity, and the functions $\Phi_1$ and $\Phi_2$ produce the tensor $\mathcal{Z}$ and $\mathcal{E}$ by merging various matrices $Z(v)$ and $E(v)$ respectively, to a three-mode tensor. It was intended to define two functions rather than one since various techniques may utilize them differently; hence, $\Phi_1$ rotates the tensor as shown in the figure \ref{fig:Constrcut_rotate_tensor}.
Tensor rotation can be advantageous in a variety of ways: Because it performs the transformation along the third dimension, the coefficients of the representation matrices are kept in the Fourier domain, reducing computing complexity greatly.

\noindent
For example, inspired by the essential connection between spectral clustering and the Markov chain, a method called essential tensor learning for multi-view spectral clustering (TLMSC) \cite{Wu_2018} builds a tensor based on multi-view transition probability matrices of the Markov chain; It first creates a similarity matrix of each view, then generates the transition probability $P^{(v)}=D^{(v)^{-1}}S^{(v)}$, and feeds it to the proposed model:
\begin{equation}
\begin{aligned}
&\min _{\mathcal{Z}, \mathcal{E}} \|\mathcal{Z}\|_{\circledast}+ \lambda\|\mathcal{E}\|_{2,1} \\
&\text { s.t. } \mathcal{P}=\mathcal{Z}+\mathcal{E}\\
&\mathcal{P}=\Phi_1\left(P^{(1)}, P^{(2)}, \ldots, P^{(V)}\right), \\
\end{aligned}
\end{equation}

\noindent where $\phi_1$ constructs and rotates the tensor to $n \times V \times n$.\\
TLMSC gives impressive result, however it employs the soft thresholding function to shrink each singular values with the same parameter. As a result, the prior knowledge of matrix singular values is not used even if it is important for the clustering, thus a method known as weighted tensor-nuclear norm minimization (WTNNM) \cite{XIE202157} is introduced to deal with this issue; It shrinks singular values with different parameters and captures the high order correlation underlying multi-view data by using $\psi$ as the weighted nuclear norm of tensor rather than the tensor nuclear norm.\\
The present tensor-nuclear norm minimization keeps some big singular values while rejecting others. This suggests that the bigger single values are significant, whilst the lower singular values are insignificant. However, in practice, such an assumption is not always correct, and higher single values may include unwanted information.\\
In \cite{Cheng2019}, a method named tensor based Representation Learning method for Multi-view Clustering (tRLMvC) \cite{Cheng2019}, adopts a different idea than the previous ones presented, it searches the representation tensor using T- product, then performs a factorization via the Tucker decomposition, the problem that is solved is:
\begin{equation}
\begin{aligned}
\min _{\mathcal{Z}, \mathcal{C}, U_1, U_3, U_3} &\|\mathcal{X}-\mathcal{X} * \mathcal{Z}\|_{F}^{2}+\alpha\|\mathcal{Z}\|_{F F 1} \\
&+\beta\left\|\mathcal{Z}-\mathcal{C} \times_{1} U_1 \times_{2} U_2 \times_{3} U_3 \right\|_{F}^{2},
\end{aligned}
\end{equation}
Then, any single view clustering algorithm such as k-means may be employed on the matrix $H=[U_1,U_2]$, with the column vector $U_3$ characterizing the contribution of each view.

\subsubsection{Tensorized graph based MVC}
To acquire clustering results, current matrix-based algorithms execute traditional spectral clustering after collecting the similarity matrix. None of them explore how to increase the stability of spectral clustering or how to utilize higher order correlation in data, hence the tensor-based technique is introduced to address this issue.

\noindent
To obtain the clustering results, the approach in \cite{Xie_2018} first pursues the subspace representation tensor using the tensor nuclear norm (different tensor rank approximations used in similar methods) before generating the final affinity matrix. Of this way, the two primary components of spectral clustering; The representation tensor and the affinity matrix, are learnt separately.

\noindent
To address this issue, a method named Graph regularized Low-rank Tensor representation and the Affinity matrix (GLTA) \cite{Chen2019} solves the problem by simultaneously learning the representation tensor and the affinity matrix, with the following optimization problem:
\begin{equation}
\begin{aligned}
\min _{\mathcal{Z}, \mathcal{E}, S, \omega} & \sum_{v=1}^{V} \Big( \alpha \operatorname{tr}\left(Z^{(v)} L^{(v)} Z^{(v)^{T}}\right)+\beta \omega_{v}\left\|Z^{(v)}-S\right\|_{F}^{2} \Big) +\|\mathcal{E}\|_{2,1}\\
&\text { s.t. } X^{(v)}=X^{(v)} Z^{(v)}+E^{(v)} \\
&\mathcal{Z}=\mathcal{C} \times{ }_{1} U_{1} \times{ }_{2} U_{2} \times{ }_{3} U_{3}, U_{i}^{\prime}U_{i}=I\\
&\mathcal{E}=\Phi_2\left(E^{(1)}, E^{(2)}, \ldots, E^{(V)}\right), \quad \omega_v \geq 0, \mathbf{1}^{T} \mathbf{\omega}=1,
\end{aligned}
\end{equation}
where $L^{(v)}$ is the normalized Laplacian matrix, GLTA learns also the local view-specific geometrical structures and the diverse contributions of distinct attributes. Specifically, when creating the final affinity matrix, it uses the Tucker decomposition to encapsulate the low-rank attribute, the manifold regularization to show the local view-specific geometrical structures, and gives appropriately varying weights to distinct characteristics.

\noindent
In the method Task-driving affinity matrix for accurate Multi-view clustering (TAMMC), \cite{Haiyan2021} tries to preserve the local affinities of all views via a graph regularization on self-expressive tensor, by penalizing a Laplacian rank on a learned common subspace by solving the following problem:
\begin{equation}
\begin{aligned}
\min _{S, \mathcal{Z}} & \sum_{v=1}^{V} \left( \alpha \operatorname{Tr}\left(Z^{(v)} L^{(v)} Z^{(v)^{T}}\right)+\lambda \|S-Z^{(v)}\|_{F}^{2} \right)\\
& + \frac{1}{2}\|\mathcal{X}-\mathcal{X} * \mathcal{Z}\|_{F}^{2}+\theta\|\mathcal{Z}\|_{T N N} \\
& \text { s.t. } \quad \mathbf{1}^{T} \mathbf{s}_{i}=1, \quad s_{i j} \geqslant 0, \operatorname{rank}\left(L_{S}\right)=n-c,
\end{aligned}
\end{equation}
where, the Laplacian matrix of each view, is calculated using Gaussian distance based on t-nearest neighbors.
The first term represents the approximation error using the Tensor Frobenius norm, the second and third terms penalize the tensor representation, thereby making it low-rank and imposing the estimated subspace to preserve the local affinity (Samples that are near in raw space are likely to remain close in the mapped subspace). The last term is used to make a consensus representation. With the rank Laplacian constraint condition, the affinity matrix is used directly for clustering.

\noindent
A more recent work, a method named Multi-view Spectral Clustering with Adaptive Graph Learning and Tensor Schatten p-norm \cite{Zhao2022}, which the suggested problem is:
\begin{equation}
\begin{aligned}
\min _{F, \mathcal{Z}, \mathbf{w}, \mathbf{\alpha}} &\sum_{v=1}^{V} \left( \frac{1}{w_{v}} \operatorname{tr}\left( X^{(v)} L^{(v)} X^{(v)} \right)+\lambda \frac{1}{\alpha_{v}} \operatorname{Tr}\left(F^{T} L^{(v)} F\right) \right) 
 +\beta\|\mathcal{Z}\|_{S p}^{p}\\
\text { s.t. } & F^{T} F=\mathbf{I}, \mathbf{1}^T \mathbf{z_{i}^{(v)}}=1, z_{i j}^{(v)} \geqslant 0 \\
&\mathbf{1}^T \mathbf{w}=1, w_{v} \geqslant 0, \mathbf{1}^T \mathbf{\alpha}=1, \alpha_{v} \geqslant 0,
\end{aligned}
\end{equation}
where $L^{(v)}$ is the Laplacian matrix of the corresponding similarity matrix $Z_{(v)}$. The Schatten p-norm regularizer is used to minimize the divergence between graphs. It guarantees also that the rank of the learned graph approximates the target rank, i.e., cluster number.\\
In practice, noises and redundancy information are frequently mixed in with the original features. As a result, the consensus similarity graph that was learned m may be erroneous. \cite{9437715} overcomes this issue by training the adaptive neighbor graph in a new low dimensional embedding space rather than the original feature space. First, it employs the view-specific similarity graph $W^{(v)}$ using an adaptive neighbor graph learning:
\begin{equation}
\begin{aligned}
\forall i=1,...,n \quad &\min_{W^{(v)}_{i}} \sum_{j=1}^{n}\left\|\mathbf{x}_{i}^{(v)}- \mathbf{x}_{j}^{(v)}\right\|_{2}^{2} w^{(v)}_{i j} 
+ \gamma w_{i j}\\
& w^{(v)}_{i j} \geq 0, \mathbf{1}^{T} \mathbf{w}^{(v)}_{i}=1.
\end{aligned}
\label{eq: CGL_step1}
\end{equation}
The new low-embedding space is learnt:
\begin{equation}
\begin{aligned}
\min_{F^{(v)}, \mathcal{T}} & -\lambda \sum_{v=1}^{V} \operatorname{tr}\left(F^{(v)} A^{(v)} F^{(v)^{T}}\right)
+\frac{1}{2}\|\mathcal{F}-\mathcal{T}\|_{F}^{2}+ \|\mathcal{T}\|_{w,*}\\
&\text { s.t. } F^{(v)^{T}} F^{(v)}=\mathbf{I}_c \\
&\mathcal{F}=[\overline{F}^{(1)} \overline{F}^{(1)^{T}};\cdots;\overline{F}^{(n)} \overline{F}^{(n)^{T}}],
\end{aligned}
\label{eq: CGL_step2}
\end{equation}
where the normalized spectral embedding matrix $\overline{F}^{(v)}$ is obtained by normalizing the rows of $F^{(v)}$, and the normalized affinity matrix is
\begin{equation}
A^{(v)}=D^{(v)^{-0.5}}W^{(v)}D^{(v)^{0.5}},
\label{eq:A}
\end{equation}
and the tensors $\mathcal{F}, \mathcal{T} \in \mathbf{R}^{n \times V \times n}$. Lastrly, the consensus graph $S$ can be learned using:
\begin{equation}
\begin{aligned}
\min_{S} & \sum_{v=1}^{V} \sum_{i, j=1}^{n}\left\|\overline{\mathbf{f}}_{i}^{(v)}- \overline{\mathbf{f}}_{j}^{(v)}\right\|_{2}^{2} s_{i j} 
+ \gamma s_{i j}\\
& s_{i j} \geq 0, \mathbf{1}^{T} \mathbf{s}_{i}=1.
\end{aligned}
\label{eq: CGL_S}
\end{equation}

\section{Experiments}\label{sec5}
In this section, we do some experiments on various data sets, with varying amounts of samples, views, and clusters. This allows for a thorough evaluation of the performance of the various algorithms. \\
For the algorithms that has $\lambda \sum_{i=1}^{c} \sigma_i(L)$ minimization part. We set $\lambda =1$, as most of the papers in their codes, in each iteration, we double $\lambda$ if the number of components is less than the number of clusters wanted, and we divide it by 2, if the number of components is less than c. In practice, we can know the comparison between the number of components and the number of clusters mentioned in \cite{Marsden2013} using the fact that if $\sum_{i=1}^{c+1} \sigma_i(L) \leq 0$ implies to the fact that number of connected components of the graph is larger than c, and $\sum_{i=1}^{c} \sigma_i(L) \geq 0$ implies that number of connected components of the graph is smaller than c.
\subsection{Data sets}
We used multiple different real world data sets used, that are commonly used in the literature, Which can be categorized into 3 Types: document, image and graph. The majority of image multi-view data sets are derived from initial single-view image data sets. The popular pre-computed multi-view picture data sets are shown. Readers can also create such multi-view data sets on their own for future research.\\
The vast majority of image multi-view data sets are created from single-view image data sets. Popular pre-computed multi-view image data sets are displayed. The document data sets are either obtained from a single view data set or naturally described in numerous views. Readers can also generate their own multi-view data sets for future investigation.
\begin{itemize}
\item \textit{One-hundred plant species leaves data set leafs (100Leaves)} \footnote{\url{https://archive.ics.uci.edu/ml/data sets/One-hundred+plant+species+leaves+data+set}}: It consists of 1600 samples, of 100 different species, 4 views are given: shape descriptor, fine scale margin and texture histogram.
\item \textit{Handwritten digit data set.} \footnote{\url{http://archive.ics.uci.edu/ml/datasets/Multiple+Features}}: is from the UCI repository. It consists of 2000 samples of handwritten digits, with a 6 views, It has 10 clusters (0-9).
\item \textit{3 sources data set (3Sources)} \footnote{\url{http://elki.dbs.ifi.lmu.de/wiki/data sets/MultiView}}: It consists of 169 news, provided by three news companies, including the BBC, Reuters, and The Guardian. Each piece of news was manually labeled with one of six relevant designations.
\item \textit{WebKB data set (WebKB)} \footnote{\url{https://linqs.soe.ucsc.edu/data}}: It consists of 203 web pages, gathered from computer science departments of university, with 4 classes, each single web page is described by the title, the content of the page and the anchor text of the hyperlink.
\item \textit{BBC data set(BBC)} \footnote{\url{http://mlg.ucd.ie/data sets/segment.html}}: It consists of 685 documents collected from the BBC news website. Each document has 4 segments with 5 manually annotated labels.
\end{itemize}
While these are provided as raw features, some graph-based approaches employ the similarity matrix as the input of each view, giving the user several options for how to generate the graph (as complete or k-nn...) and the metrics( as binary,cosine, or gaussian kernel...) from the raw input features. We generate exhaustive graphs from the same data set as previously, and we display the best performance on the best graph. We also included two pre-computed graph data sets:
\begin{itemize}
\item \textit{MSRC-v1} \footnote{\url{http://mldta.com/dataset/msrc-v1/}} is a class-based picture collection from Microsoft Research in Cambridge containing 240 samples, with 7 classes. Each picture is represented by 6 distinct features: the HOG, the Color moment, the CENT, the LBP, the CENTRIST, and the SIFT feature.
\item \textit{Yale Face Database(Yale)} \footnote{\url{https://www.robots.ox.ac.uk/~vgg/data/flowers/17/}}
The database contains 165 GIF images with 15 classes. There are 11 ²s per subject.
\item \textit{17 Category Flower Dataset(Flo17)} \footnote{\url{www.robots.ox.ac.uk/vgg/data/flowers/17}} is a class-based picture of 17 type of flowers, with 80 images of each class, which means, a total of 1360 samples. It has 17 distinct features.
\end{itemize}
The characteristics of these raw feature data sets are given in the table below \ref{Tab:data_sets}, which displays the number of samples, views, clusters, and $d_v$, which represents the dimension of features in view $v$. More information on the data set may be obtained by following the hyperlinks in the footnotes.
\begin{table}[ht]
\begin{minipage}{\textwidth}
\caption{Summary of the benchmark data sets used}
\label{Tab:data_sets}
\centering
\begin{tabular}{lccccccccc}
\hline Data set & \# samples & \# views & \# clusters & $\mathrm{d}_1$ & $d_2$ & $d_3$ & $d_4$ & $d_5$ & $d_6$ \\
\hline
3Sources & 169 & 3 & 6 & 3560 & 3631 & 3068 & - & - & - \\
WebKB & 203 & 3 & 4 & 1703 & 230 & 230 & - & - & - \\
100Leaves & 1600 & 3 & 100 & 64 & 64 & 64 & - & - & - \\
BBC & 685 & 4 & 5 & 4659 & 4633 & 4665 & 4684 & - & - \\
HW & 2000 & 6 & 10 & 216 & 76 & 64 & 6 & 240 & 47 \\
\hline
\end{tabular}
\end{minipage}
\end{table}
\subsection{Compared methods}
Following that, we will compare the state of the art algorithms, for which public code, A couple modification were done on these data set in order to work with the same data set, and to work with the same clustering measures. The k-means algorithm adopted is also available. Note that the code of some algorithms were created from scratch, since they were not available to the public, but we majorly focused on the methods that the code exist in any programming language. The different methods are summarized in the table below, where we've devised the methods on two columns: Methods using the matrix related theory, the ones used the tensor related theory. We distinguish by emphasizing the methods that require the similarity matrices instead as input.
\begin{table}[ht]
\begin{center}
\begin{minipage}{200pt}
 \caption{Benchmark methods}
\begin{tabular}{c c |c }
\hline
\multicolumn{2}{c}{\textbf{Matricized}} 
& \textbf{Tensorial} \\ \hline
GBS\tablefootnote{\url{https://github.com/cswanghao/gbs}}
& CSMSC\tablefootnote{\url{https://github.com/XIAOCHUN-CAS/Consistent-and-Specific-Multi-View-Subspace-Clustering}}
& t-SVD-MSC\tablefootnote{\url{https://www.researchgate.net/publication/324151918_the_source_of_paper_On_Unifying_Multi-View_Self-Representations_for_Clustering_by_Tensor_Multi-Rank_Minimization}} \\

GMC\tablefootnote{\url{https://github.com/cshaowang/gmc}}
& MVLRSCC\tablefootnote{\url{https://github.com/mbrbic/Multi-view-LRSSC}}
& MVSC-TPR\tablefootnote{\url{http://www.scholat.com/portaldownloadFile.html?fileId=4623}} \\

MCGC\tablefootnote{\url{https://github.com/kunzhan/MCGC}}
& MVGL\tablefootnote{\url{https://github.com/kunzhan/MVGL}}
& CGL\tablefootnote{\url{https://github.com/guanyuezhen/CGL}} \\

AMGL\tablefootnote{\url{https://github.com/kylejingli/AMGL-IJCAI16}}
& MLAN\tablefootnote{\url{https://github.com/jxqhhh/MLAN}}
& \underline{\textit{MVSC-TLRR}}\tablefootnote{\url{https://github.com/jyh-learning/MVSC-TLRR}} \\

\underline{\textit{SwMC}}\tablefootnote{\url{https://github.com/kylejingli/SwMC-IJCAI17}}
& \underline{\textit{RG-MVC}}\tablefootnote{\url{https://github.com/wx-liang/RG-MVC}}
& \underline{\textit{MCSCS p\_norm}}\tablefootnote{Created from algorithm in the paper \cite{Zhao2022}} \\ \hline

\end{tabular}
\end{minipage}
\end{center}
\label{Tab:methods}
\end{table}

\subsection{Evaluation metrics}
Each algorithm is run ten times to randomize the experiments. Different measures are utilized to evaluate the performance of these multi view clustering algorithms based using well-known benchmark data sets. In the clustering task, several measures favor various traits. \\
The accuracy (ACC), the normalized mutual information (NMI), the F1 measure (F-measure), and the adjusted rand index (ARI) are the main ones used:

\noindent
- NMI \citep{Strehl2002} is an information theoretic metric that measures the mutual information between the cluster assignments and the ground truth labels. It is normalized by the average of entropy of both ground labels and the cluster assignments, it even allows the comparison of two partitions even with a different number of clusters:
\begin{equation}
NMI=\dfrac{2I(\Omega,\Theta)}{H(\Omega)H(\Theta)},
\end{equation}
where $I$ represents the mutual information, and $H$ the information entropy.\\
$\Omega = \{ \omega_1, \omega_2, \ldots, \omega_K \}$ is the set of clusters and $\Theta = \{ c_1,c_2,\ldots,c_J \}$ is the set of classes.\\
-ARI \citep{Hubert1985} computes a similarity measure between two clustering by considering all pairs of samples and counting pairs that are assigned in the same or different clusters in the predicted and true clustering. The adjusted Rand index is the corrected-for-chance version of the Rand index. It also allows the comparison of two partitions with different number of clusters: 
\begin{equation}
\begin{aligned}
ARI & =\dfrac{RI-Expected(RI)}{max(RI)-Expected(RI)}\\
& =\dfrac {\sum _{ij}{\binom {n_{ij}}{2}}-\frac{1}{\binom {n}{2}} \sum _{i}{\binom {n_{i}}{2}}\sum _{j}{\binom {n_{j}}{2}}}{{\frac {1}{2}}\left[\sum _{i}{\binom {n_{i}}{2}}+\sum _{j}{\binom {n_{j}}{2}}\right]-\frac{1}{\binom {n}{2}} \sum _{i}{\binom {n_{i}}{2}}\sum _{j}{\binom {n_{j}}{2}}},
\end{aligned}
\end{equation}
where $n$ represents the number of samples, $n_i= \# c_i$ is the number of samples in the cluster $c_i$, while $n_j= \# w_j$ is the number of samples in the class $w_j$ and $n_{i}^j =\# \lvert w_i \cap c_j \rvert $ is the number of data points of cluster $w_i$ assigned to the class $c_j$. \\
In general, an ARI value lies between 0 and 1. The index value is equal to 1 only if a partition is completely identical to the intrinsic structure and close to 0 for a random partition.\\
-Purity\citep{Manning_2008} is a simple and transparent evaluation of the degree of each cluster containing a class of data points. It's value can be calculated via the following formula:
\begin{equation}
Purity=\frac{1} {n} \sum_i \max_j n_{i}^j.
\end{equation}
-ACC \citep{Laszlo1986} is the unsupervised counterpart of classification accuracy. It differs from usual accuracy metrics in that it employs a mapping function $map$ to determine the optimal mapping between the algorithm's cluster assignment output $Y$ and the ground truth $Y^0$. This mapping is essential because an unsupervised algorithm may use a label other than the actual ground truth label to describe the same cluster, The best mapping can be found by using the Kuhn–Munkres algorithm in \cite{Laszlo1986}:
\begin{equation}
ACC=\max_{map} \dfrac{\sum_{i=1}^{n} \delta(Y_i^0, map(Y_i))}{n},
\end{equation}
where $\delta(a,b)$ is a judgment function, it gives 1 if $a=b$ and 0 otherwise.\\
-$F_\beta$ measure is another metric that will be used, that can be calculated from the recall $R$ and the precision $P$ of a test, there is another parameter $\beta$( where it will be used equals to 1 in the numerical tests), which gives us the control of penalizing the false negatives more strongly than false positives by selecting a value greater than 1, as in ACC, we also used the permutation mapping that maximizes the matching of ground truth and predictive labels: 
\begin{eqnarray*}
P = \frac{TP}{TP+FP} \qquad
R = \frac{TP}{TP+FN} \qquad
F_{\beta} = \frac{(\beta^2+1)PR}{\beta^2 P+R},
\end{eqnarray*}
where $TP,TN,FP$, and $FN$ denote the number of similar pairs (same class) assigned to the correct cluster, number of dissimilar pairs (different class) assigned to the wrong correct cluster, number of dissimilar pairs assigned to the wrong cluster, and number of similar pairs assigned to the wrong cluster, respectively.\\
For all metrics, a higher value indicates the better performance of the clustering.
\subsection{Comparison result and analysis}
For some methods, a post clustering (k -means) is needed on the obtaining embedding data, as it is sensitive to initial values, we repeat the k-means clustering processing 20 trials to avoid the randomness perturbation and report the result with the lowest value for the objective function of k-means clustering among the 20 results.\\
The comparison results are shown in Tables \ref{Tab:hw_measures}, \ref{Tab:bbc_measures}, \ref{Tab:WebKB_measures}, \ref{Tab:3sources_measures}, and \ref{Tab:leaves_measures}. The best value of each measure is shown in bold. We note that the standard deviation is avoided to be show for the k-means clustering trials since simply, not all the methods use it, we also mention that the values are rounded to the ten-thousandths:

\begin{table}[H]
 \centering
 \begin{minipage}{\textwidth} 
 \caption{Clustering measure on 3sources data set}
\label{Tab:3sources_measures}
\resizebox{\columnwidth}{!}{%
 \begin{tabular}{l c c c c c c c c }
 \hline
 \textbf{Method} & \textbf{Fscore} & \textbf{Precision} & \textbf{Recall} & \textbf{NMI} & \textbf{ARI} & \textbf{ACC} & \textbf{Purity} \\ \hline
\textbf{CGL} & 0.6224 & 0.7097 & 0.5555 & 0.6800 & 0.5263 & 0.6604 & 0.7976 \\ 
\textbf{GBS} & 0.7159 & 0.6052 & 0.8760 & 0.7096 & 0.6082 & 0.7692 & 0.8047 \\ 
\textbf{GMC} & 0.6047 & 0.4844 & 0.8045 & 0.6216 & 0.4431 & 0.6923 & 0.7456 \\ 
\textbf{AMGL } & 0.2765 & 0.1711 & 0.7604 & \textbf{0.8490} & 0.2652 & 0.6352 & 0.7028 \\ 
\textbf{T-SVD-MS} & 0.7029 & 0.7100 & 0.6961 & 0.6362 & 0.6098 & 0.7547 & 0.7967 \\ 
\textbf{MVSC-TPR} & 0.5420 & 0.5722 & 0.5172 & 0.4561 & 0.4117 & 0.5979 & 0.6757 \\ 
\textbf{CSMSC} & 0.3800 & \textbf{1.0000} & 0.2346 & 0 & 0 & 0.3299 & \textbf{1.0000} \\ 
\textbf{MVGL } & 0.3455 & 0.2255 & 0.7381 & 0.1522 & 0.0163 & 0.3846 & 0.4320 \\ 
\textbf{Pairwise MLRSSC} & 0.6443 & 0.6045 & 0.6969 & 0.6167 & 0.5460 & 0.6840 & 0.7491 \\ 
\textbf{Centroid MLRSSC} & 0.5662 & 0.4629 & 0.7492 & 0.6470 & 0.4725 & 0.6482 & 0.6683 \\ 
\textbf{PairwiseKMLRSSC} & 0.4675 & 0.3798 & 0.6118 & 0.4837 & 0.3525 & 0.5098 & 0.5240 \\ 
\textbf{Centroid KMLRSSC} & 0.4745 & 0.3847 & 0.6214 & 0.4824 & 0.3620 & 0.5169 & 0.5272 \\ 
\textbf{MLAN} & \textbf{0.9112} & 0.9112 & \textbf{ 0.9112} & 0.7881 & \textbf{0.8171} & \textbf{0.9112 } & 0.9112 \\ 
 \hline
 \end{tabular}
 }
 \end{minipage}
\end{table}

\begin{table}[H]
 \centering
\begin{minipage}{\textwidth} 
 \caption{Clustering measure on WebKB data set}
 \label{Tab:WebKB_measures}
\resizebox{\columnwidth}{!}{%
 \begin{tabular}{l c c c c c c c c }
 \hline
 \textbf{Method} & \textbf{Fscore} & \textbf{Precision} & \textbf{Recall} & \textbf{NMI} & \textbf{ARI} & \textbf{ACC} & \textbf{Purity } \\ \hline
 \textbf{CGL} & 0,5025 & \textbf{0,6249 }& 0,4202 & 0,3178 & 0,2725 & 0,4729 & 0,7783 \\ 
 \textbf{GBS} & \textbf{0,7227} & 0,6148 & 0,8766 & \textbf{0,465} &\textbf{ 0,4844} &\textbf{ 0,7833} &\textbf{ 0,798} \\ 
 \textbf{GMC} & 0,6878 & 0,5821 & 0,8404 & 0,4133 & 0,4169 & 0,7586 & 0,7734 \\ 
 \textbf{AMGL} & 0,5489 & 0,3925 &\textbf{ 0,9128} & 0,037 & -0,002 & 0,5099 & 0,5308 \\ 
 \textbf{T-SVD-MS} & 0,562 & 0,5642 & 0,5609 & 0,2536 & 0,2772 & 0,6182 & 0,6805 \\ 
 \textbf{MVSC-TPR} & 0,4751 & 0,5664 & 0,4092 & 0,2042 & 0,217 & 0,5539 & 0,6941 \\ 
 \textbf{CSMSC} & 0,4533 & 0,2435 & 0,509 & 0,4755 & 0,2983 & 0,4102 & 0,4835 \\ 
 \textbf{MVGL} & 0,6144 & 0,5761 & 0,6581 & 0,2805 & 0,3358 & 0,6995 & 0,7389 \\ 

 \textbf{Pairwise MLRSSC} & 0,4453 & 0,3339 & 0,6746 & 0,3675 & 0,2504 & 0,4204 & 0,4493 \\ 
 \textbf{Centroid MLRSSC} & 0,3816 & 0,2629 & 0,7036 & 0,3539 & 0,214 & 0,3963 & 0,4032 \\ 
 \textbf{Pairwise KMLRSSC} & 0,4655 & 0,3818 & 0,6021 & 0,2438 & 0,2338 & 0,5234 & 0,57 \\ 
 \textbf{Centroid KMLRSSC} & 0.4528 & 0.3649 & 0.6043 & 0.237 & 0.2264 & 0.5015 & 0.5436 \\ 
 \textbf{MLAN} & 0.3448 & 0.7881 & 0.2226 & 0.4921 & 0.5119 & 0.7881 & 0.8029 \\ \hline
 \end{tabular}
 }
\end{minipage}
\end{table}

\begin{table}[H]
 \centering
 \begin{minipage}{\textwidth} 
 \caption{Clustering measure on 100Leaves data set}
 \label{Tab:leaves_measures}
\resizebox{\columnwidth}{!}{%
 \begin{tabular}{l c c c c c c c c }
 \hline
 \textbf{Method} & \textbf{Fscore} & \textbf{Precision} & \textbf{Recall} & \textbf{NMI} & \textbf{ARI} & \textbf{ACC} & \textbf{Purity} \\ \hline
 \textbf{CGL} & 0,6245 & 0,7097 & 0,5654 & 0,6871 & 0,5263 & 0,6604 & 0,8024 \\
 \textbf{GBS} & 0,3867 & 0,2555 & 0,7951 & 0,8711 & 0,3779 & 0,6694 & 0,7131 \\
 \textbf{GMC} & 0,5042 & 0,3521 & \textbf{0,8874} & 0,9292 & 0,4974 & 0,8237 & 0,8506 \\
 \textbf{AMGL} & 0,2765 & 0,1711 & 0,7604 & 0,849 & 0,2652 & 0,6352 & 0,7028 \\
 \textbf{T-SVD-MS} & \textbf{0,8549} & 0,9428 & 0,7825 & \textbf{0,9708} & \textbf{0,8534} & \textbf{0,8568 }& 0.9606\\ 
 \textbf{MVSC-TPR} & 0,5787 & 0,5209 & 0,6515 & 0,8532 & 0,5743 & 0,6607 & 0,7028 \\
 \textbf{CSMSC} & 0,3455 & 0,5543 & 0,4521 & 0.6433 & 0,3894 & 0,3467 & 0,4870 \\
 \textbf{MVGL} & 0,2299 & 0,1545 & 0,4491 &\textbf{ 0,732} & 0,219 & 0,5056 & 0,5463 \\

 \textbf{Pairwise MLRSSC} & 0,0186 &\textbf{ 0,9994} & 0,0094 & 0,0006 & 0 & 0,0103 & \textbf{0,9997 }\\
 \textbf{Centroid MLRSSC} & 0,0186 & 0,9991 & 0,0094 & 0,0009 & 0,000 & 0,0104 & 0,9995 \\
 \textbf{Pairwise KMLRSSC} & 0,0824 & 0,6515 & 0,044 & 0,4501 & 0,066 & 0,0781 & 0,7601 \\ 
 \textbf{Centroid KMLRSSC} & 0.4528 & 0.0821 & 0.663 & 0.0438 & 0.4523 & 0.0656 & 0.0784\\ 
 \textbf{MLAN} & 0.0706 & 0.0706 & 0.0706 & 0.9494 & 0.8242 & 0.87375 & 0.8956 \\ \hline
 \end{tabular}
 }
 \end{minipage}
\end{table}

\begin{table}[H]
 \centering
 \begin{minipage}{\textwidth} 
 \caption{Clustering measure on BBC data set}
 \label{Tab:bbc_measures}
 \resizebox{\columnwidth}{!}{%
 \begin{tabular}{l c c c c c c c c }
 \hline
\textbf{Method} & \textbf{Fscore} & \textbf{Precision} & \textbf{Recall} & \textbf{NMI} & \textbf{ARI} & \textbf{ACC} & \textbf{Purity} \\ \hline
\textbf{CGL} & \textbf{0,9061} & \textbf{0,8906} & \textbf{0,9269} & \textbf{0,9185} &\textbf{ 0,8761} & \textbf{0,8886} &\textbf{ 0,9296} \\ 
\textbf{GBS} & 0,5451 & 0,4031 & 0,8414 & 0,4839 & 0,3337 & 0,6044 & 0,6102 \\ 
\textbf{GMC} & 0,6333 & 0,5012 & 0,86 & 0,5628 & 0,4789 & 0,6934 & 0,6934 \\ 
\textbf{AMGL} & 0,3748 & 0,238 & 0,8891 & 0,02 & 0,0082 & 0,3431 & 0,3477 \\ 
\textbf{T-SVD-MS} & 0,3516 & 0,7055 & 0,2361 & 0,0896 & 0,0034 & 0,3396 & 0,8132 \\ 
\textbf{MVSC-TPR} & 0,574 & 0,6430 & 0,4039 & 0,3498 & 0,2344 & 0,4366 & 0,8132 \\ 
\textbf{CSMSC} & 0,48374 & 0,5903 & 0,6345 & 0,243 & 0,245 & 0,245 & 0,674 \\ 
\textbf{MVGL} & 0,3686 & 0,2329 & 0,8841 & 0,0806 & 0,0043 & 0,3489 & 0,3509 \\ 
\textbf{Pairwise MLRSSC} & 0,7291 & 0,6643 & 0,8118 & 0,7213 & 0,6571 & 0,7302 & 0,7673 \\ 
\textbf{Centroid MLRSSC} & 0,6498 & 0,5202 & 0,8669 & 0,6971 & 0,5751 & 0,6562 & 0,6562 \\ 
\textbf{Pairwise KMLRSSC} & 0,408 & 0,3338 & 0,5274 & 0,4228 & 0,2761 & 0,4622 & 0,4725 \\ 
\textbf{Centroid KMLRSSC} &0.4051 & 0.3442 & 0.4948 & 0.4093 & 0.263 & 0.4661 & 0.4885 \\ 
\textbf{MLAN } & 0.5591 & 0.5591 & 0.5591 & 0.7274 & 0.7460 & 0.8715 & 0.8715 \\ 
\hline
 \end{tabular}
 }
 \end{minipage}
\end{table}


\begin{table}[H]
 \centering
\begin{minipage}{\textwidth} 
 \caption{Clustering measure on HW data set}
 \label{Tab:hw_measures}
 \resizebox{\columnwidth}{!}{%
 \begin{tabular}{l c c c c c c c c }
 \hline
\textbf{Method} & \textbf{Fscore} & \textbf{Precision} & \textbf{Recall} & \textbf{NMI} & \textbf{ARI} & \textbf{ACC} & \textbf{Purity } \\ \hline
\textbf{CGL} & 0,8584 & 0,7941 & 0,963 &0,9354 & 0,8405 & 0,8617 & 0,8741 \\ 
\textbf{GBS} & 0,8327 & 0,7788 & 0,8945 & 0,8787 & 0,8128 & 0,847 & 0,8655 \\ 
\textbf{GMC} & 0,8653 & 0,826 & 0,9085 & 0,9041 & 0,8496 & 0,882 & 0,882 \\ 
\textbf{AMGL} & 0,4011 & 0,3081 & 0,5788 & 0,6162 & 0,3115 & 0,5536 & 0,5893 \\ 
\textbf{T-SVD-MS} & \textbf{0,9042 }&\textbf{ 0,9729} & \textbf{0,8464} & 0,9602 & \textbf{0,8926 }&\textbf{ 0,8936 }& \textbf{0,983} \\ 
\textbf{MVSC-TPR} & 0,6745 & 0,6346 & 0,8045 & 0,8341 & 0,7563 & 0,7732 & 0,7422 \\ 
\textbf{CSMSC} & 0,4354 & 0,6349 & 0,7409 & 0,753 & 0,65 & 0,632 & 0,6244 \\ 
\textbf{MVGL} & 0,6367 & 0,557 & 0,7429 & 0,7231 & 0,59 & 0,6895 & 0,7245 \\ 
\textbf{Pairwise MLRSSC} & 0,6498 & 0,5202 & 0,8669 & 0,6971 & 0,5751 & 0,6562 & 0,6562 \\ 
\textbf{Centroid MLRSSC} & 0,7266 & 0,8678 & 0,625 & 0,8021 & 0,6907 & 0,7438 & 0,9244 \\ 
\textbf{Pairwise KMLRSSC} & 0,5399 & 0,6179 & 0,4794 & 0,5958 & 0,4818 & 0,6361 & 0,7412 \\ 
\textbf{Centroid KMLRSSC} & 0.5399 & 0.6179 & 0.4794 & 0.5958 & 0.4818 & 0.6361 & 0.7412 \\
\textbf{MLAN} & 0.1965 & 0.1965 & 0.1965 & 0.\textbf{9374} & 0.9384 & 0.972 & 0.972 \\ \hline
 \end{tabular}
 }
\end{minipage}
\end{table}

\noindent
As it shows from the tables, there is clearly, an advantage of using the tonsorial theory, these type methods has a batter performances on all kind of measures on most of the data. 

\begin{table}[H]
 \centering
 \begin{minipage}{\textwidth} 
 \caption{Clustering measure of graph methods on different data set}
 \label{Tab:Graphs_measures}
 \resizebox{\columnwidth}{!}{%
 \begin{tabular}{l l c c c c c c c c}
 \hline
 \textbf{Method} & \textbf{Data set} & \textbf{Fscore} & \textbf{Precision} & \textbf{Recall} & \textbf{NMI} & \textbf{ARI} & \textbf{ACC} & \textbf{Purity } \\ \hline
 \textbf{SwMC} & 3sources & \textbf{0.5888} & 0.4637 & \textbf{0.8063} & 0.6036 & \textbf{0.4166} & \textbf{ 0.6746 }& 0.7278 \\ 
 \textbf{RG-MVC} & 3sources & 0.4956 & 0.456 & 0.7963 &\textbf{ 0.6354 }& 0.345 & 0.5244 & 0.6833 \\ 
 \textbf{MCSCS p\_norm} & 3sources & 0.2768 & 0.3422 & 0.2324 & 0 & 0 & 0.3314 & 0.3442 \\ 
 \textbf{MVSC-TLRR} & 3sources & \textbf{0.5197 }& 0.7299 & 0.4138 & 0.4498 & 0.3167 & 0.5707 & \textbf{0.8272} \\ \hline
 \textbf{SwMC} & BBC & \textbf{0.6167} & 0.4846 &\textbf{ 0.8476} & 0.5534 & \textbf{0.4536} & \textbf{0.6964} & 0.6964 \\ 
 \textbf{RG-MVC} & BBC & 0.3453 & 0.3948 & 0.7340 & 0.345 & 0.354 & 0.5464 & 0.345 \\ 
 \textbf{MCSCS p\_norm} & BBC & 0.38 & 0.439 & 0.2346 & 0 & 0 & 0.3299 & 0.3847 \\ 
 \textbf{MVSC-TLRR} & BBC & 0.564 & 0.4566 & 0.4353 & \textbf{0.6309} & 0.3453 & 0.5635 &\textbf{ 0.7353} \\ \hline
 \textbf{SwMC} & Flower & 0.2804 & 0.181 & 0.6221 & 0.5566 & 0.1506 & 0.5391 & 0.53 \\ 
\textbf{RG-MVC} & Flower & 0.1621 & 0.1834 & 0.1453 & 0.2342 & 0.1931 & 0.3453 & 0.2431 \\ 
\textbf{MCSCS p\_norm} & Flower & 0.1773 & 0.1917 & 0.1654 & 0.2886 & 0.1225 & 0.2709 & 0.319 \\ 
 \textbf{MVSC-TLRR} & Flower & 0.2666 & 0.2726 & 0.2609 & 0.4108 & 0.2203 & 0.3781 & \textbf{0.4082} \\ \hline
 \textbf{SwMC} & HW & 0.7668 & 0.6849 & \textbf{0.8711} & 0.862 & \textbf{0.7376} & \textbf{0.75} & \textbf{0.787 }\\ 
 \textbf{RG-MVC} & HW & 0.6734 & 0.435 & 0.6543 & 0.739 & 0.945 & 0.6453 & 0.6953 \\
 \textbf{MCSCS p\_norm} & HW & 0.5026 & 0.5733 & 0.4505 & 0.5886 & 0.44 & 0.5969 & 0.6924 \\ 
 \textbf{MVSC-TLRR} & HW & 0.\textbf{7745} & \textbf{0.6934} & 0.8313 & \textbf{0.876} & 0.735 & 0.654 & 0.6941 \\ \hline
 \textbf{SwMC} & Leaves & 0.3644 & 0.2949 & 0.4766 & 0.7625 & 0.3569 & 0.5581 & 0.5844 \\
 \textbf{RG-MVC} & Leaves & 0.2409 & 0.2584 & 0.4209 & 0.7433 & 0.3583 & 0.5143 & 0.5243 \\
 \textbf{MCSCS p\_norm} & Leaves & \textbf{0.4079} & 0.4743 & 0.3582 & 0.7605 & 0.4015 & 0.5502 & 0.6264 \\ 
 \textbf{MVSC-TLRR} & Leaves & 0.2138 & 0.3646 & 0.1517 & 0.6735 & 0.2032 & 0.4077 & 0.5272 \\ \hline
 \textbf{SwMC} & MSRCV1 & \textbf{0.8412} & \textbf{0.8304} & \textbf{0.8522} & \textbf{0.853} & \textbf{0.8152 }&\textbf{ 0.919} &\textbf{ 0.919} \\ 
 \textbf{RG-MVC} & MSRCV1 & 0.4946 & 0.5314 & 0.4632 & 0.6326 & 0.4611 & 0.5855 & 0.6778 \\ 
\textbf{MCSCS p\_norm} & MSRCV1 & 0.3856 & 0.3314 & 0.4612 & 0.5936 & 0.3947 & 0.4634 & 0.4563 \\
 \textbf{MVSC-TLRR} & MSRCV1 & 0.7368 & 0.7988 & 0.6889 & 0.7824 & 0.6896 & 0.785 & 0.8667 \\ \hline
 
 \textbf{SwMC} & WebKB & \textbf{0.7004} & \textbf{0.5919} &\textbf{ 0.8576} &\textbf{ 0.4351} & \textbf{0.4397} & \textbf{0.7685 } & \textbf{0.7833} \\ 
 \textbf{RG-MVC} & WebKB & 0.6433 & 0.5948 & 0.7003 & 0.4112 & 0.534 & 0.493 & 0.7103 \\ 
 \textbf{MCSCS p\_norm} & WebKB & 0.4945 & 0.4347 & 0.5734 & 0 & 0 & 0.5271 & 0.593 \\
 \textbf{MVSC-TLRR} & WebKB & 0.6944 & 0.5903 & 0.8344 & 0.4230 & 0.4093 & 0.6734 & 0.7304 \\ \hline
 \textbf{SwMC} & Yale & \textbf{0.7734} & 0.6343 & 0.6301 & 0.6781 & 0.6433 & 0.823 & 0.8423 \\ 
 \textbf{RG-MVC} & Yale & 0.6434 & 0.6121 & 0.5834 & 0.6453 & 0.5498 & 0.7 & 0.7994 \\
 \textbf{MCSCS p\_norm} & Yale & 0.5759 & 0.6012 & 0.5529 & 0.6207 & 0.5041 & 0.6852 & 0.7212 \\
 \textbf{MVSC-TLRR} & Yale & 0.7519 & \textbf{0.8386} & \textbf{0.6875} & \textbf{0.8278} & \textbf{0.7209 }& \textbf{0.7841} & \textbf{0.9084} \\ \hline
 \end{tabular}
 }
\end{minipage}
\end{table}

\subsection{Further work}
We can also, compare the time complexity, but since the methods are not all in the same language (Matlab and Python), we decided not to stop in order to not falsify the reel comparison between the methods.\\
For the graph based methods, we can further investigate the hyper graph, which differs from the graph by letting the edges to be associated to more than two nodes, as we can calculate the Laplacian by an appropriate formula, and then continue the rest of the word as the normal graphs, as in \cite{Lu2020}, named HLR-MSCLRT, where the following objective is the same as in \eqref{eq:Subspace tensor MVC}, by simply replacing the Laplacian of a graph matrix of a view $L^{(v)}$ by its Hyper graph Laplacian $L_{h}^{(v)}$.

\section{Conclusion}\label{sec6}
As it is shown, for clustering multi view is the way to go, as it brings with its multiple views, multiple data that can give more information, having also new methods, a part from the ones that can be generalized from the single view, without forgetting the new challenges of how to combine and use all the views, the tensor representation of the data, has a clear advantages of just considering data as matrices, by exploiting the higher order correlation in the data, which has been used, different norms and ranks, The coherence and complimentary nature of diverse views of view are therefore automatically taken into account.
Two main categories were mainly discussed are the graph based, and the subspace based, we can even give the graph based method more credits, as it gives insight on the local view specific geometrical structures, and can also be combined in some methods that tries to find a low rank representation tensor with a Laplacian term, while also some method tries to learn the affinity matrices of each view and others only the global affinity matrix.

\bibliographystyle{unsrtnat}


\end{document}